%% file: template.tex
\documentclass[cameraready]{Interspeech}
\usepackage{booktabs}
\usepackage{tipa}
\usepackage{booktabs}
\usepackage{tabularx}
\usepackage[table]{xcolor}

\usepackage{comment}
\definecolor{darkgreen}{RGB}{0,100,0} 
\hypersetup{
    colorlinks=true,
    linkcolor=red,
    urlcolor=purple,
    citecolor=blue
}

\title{
Bolbosh: Script-Aware Flow Matching for Kashmiri Text-to-Speech
}

\author[affiliation={1,3}, correspondingauthor]{Tajamul}{Ashraf}

\author[affiliation={3}, equalcontribution]{Burhaan Rasheed}{Zargar}

\author[affiliation={3}, equalcontribution]{Saeed Abdul}{Muizz}

\author[affiliation={2}]{Ifrah}{Mushtaq}

\author[affiliation={2}]{Nazima}{Mehdi}

\author[affiliation={3}]{Iqra Altaf}{Gillani}

\author[affiliation={2}]{Aadil Amin}{Kak}

\author[affiliation={3}]{Janibul}{Bashir}

\address{
   $^1$ King Abdullah University of Science and Technology (KAUST)
   $^2$ Department of Linguistics, University of Kashmir.
    $^3$ Gaash Lab, National Institute of Technology Srinagar 
}

\email{https://gaash-lab.github.io/Bolbosh/}


\begin{document}
\maketitle

\begin{abstract}
Kashmiri is spoken by around 7 million people, but remains critically underserved in speech technology. despite its official status and rich linguistic heritage. The lack of robust Text-to-Speech (\texttt{TTS}) systems limits digital accessibility and inclusive human-computer interaction for native speakers. In this work, we present the first dedicated, open-source neural \texttt{TTS} system designed for Kashmiri. We show that zero-shot multilingual baselines trained for Indic languages fail to produce intelligible speech, achieving a Mean Opinion Score (MOS) of only 1.86, largely due to inadequate modeling of Perso-Arabic diacritics and language-specific phonotactics.
To address these limitations, we propose \textbf{Bolbosh}, a supervised cross-lingual adaptation strategy based on Optimal Transport Conditional Flow Matching (OT-CFM) within the Matcha-TTS framework. This enables stable alignment under limited paired data. We further introduce a three-stage acoustic enhancement pipeline consisting of dereverberation, silence trimming, and loudness normalization to unify heterogeneous speech sources and stabilize alignment learning. The models's vocabulary is expanded to explicitly encode Kashmiri graphemes, preserving fine-grained vowel distinctions.
Our system achieves a MOS of 3.63 and a Mel-Cepstral Distortion (MCD) of 3.73, substantially outperforming multilingual baselines and establishing a new benchmark for Kashmiri speech synthesis. Our results demonstrate that script-aware and supervised flow-based adaptation are critical for low-resource TTS in diacritic-sensitive languages. Code and data are available at \href{https://github.com/gaash-lab/Bolbosh}{https://github.com/gaash-lab/Bolbosh}
\end{abstract}

\input{sections/introduction}
\input{sections/method}
\input{sections/results}

\input{sections/conclusion}

\bibliographystyle{IEEEtran}
\bibliography{mybib}

\end{document}

%% file: sections/introduction.tex
\section{Introduction}
\label{sec:intro}

Kashmiri is a Dardic language within the Indo-Aryan branch of the Indo-European family \cite{wali1997kashmiri, koul1987spoken, abdullah2025endangered, wade1995grammar, qumar2024addressing}, spoken by approximately 7 million people in the Kashmir Valley and diaspora communities \cite{census2011kashmiri}. Beyond its sociolinguistic significance, Kashmiri has attracted sustained global academic interest due to its typologically distinctive Indo-Aryan features and its historical blending of Central Asian and Indian linguistic influences. Despite its official status and rich literary heritage, Kashmiri remains severely under-resourced in speech technology. The absence of high-quality Text-to-Speech (TTS) systems limits accessibility, digital participation, and inclusive human-computer interaction, widening the digital divide as voice-driven interfaces become increasingly prevalent.

Developing TTS for Kashmiri poses challenges typical of low-resource languages, compounded by linguistic complexity. Paired text–speech corpora are scarce and fragmented, limiting the scalability of neural models. Kashmiri is written in multiple scripts, primarily Perso-Arabic, Devanagari and Roman, creating orthographic inconsistencies; the Perso-Arabic script heavily relies on diacritics that encode subtle vowel distinctions essential for intelligibility. Additionally, dialectal variation in lexicon, phonology, and prosody complicates alignment and generalization. Together, these factors can substantially degrade zero-shot multilingual TTS performance for Kashmiri.

\noindent\textbf{Limitations of Existing Multilingual Baselines.}
Recent multilingual \texttt{TTS} frameworks such as IndicParler \cite{sankar25_interspeech} have expanded coverage across Indic languages with promising zero-shot performance. However, Kashmiri remains unofficially supported and lacks supervised adaptation. Our evaluation shows that these models perform poorly on Kashmiri, achieving a Mean Opinion Score (MOS) of only 1.86, with frequent vowel mispronunciations, prosodic distortion, and reduced intelligibility. We attribute this to inadequate modeling of Perso-Arabic diacritics and phonotactic mismatches with high-resource training languages, indicating that zero-shot multilingual transfer is insufficient for preserving Kashmiri’s acoustic integrity.

\noindent\textbf{Proposed Approach: OT-CFM for Low-Resource Adaptation.}
To address these limitations, we propose \textbf{Bolbosh}, a supervised adaptation strategy based on Optimal Transport Conditional Flow Matching (OT-CFM) \cite{tong2023conditional} within the Matcha-TTS architecture. \textbf{Bolbosh} models speech generation as a continuous transport from a Gaussian before the target acoustic distribution, which promotes stable monotonic alignment and improved sample efficiency under limited paired data, enabling stable alignment and efficient training.
We initialize from a pretrained multi-speaker English checkpoint and fine-tune on a curated 79.9-hour Kashmiri corpus combining studio-quality RASA \cite{ai4bharat2024rasa} and spontaneous IndicVoices-R data \cite{sankar2024indicvoicesr}. To mitigate domain mismatch, we apply a three-stage enhancement pipeline: dereverberation, dynamic silence trimming, and LUFS normalization.
Finally, we expand the model's vocabulary to 272 graphemes to include support for Kashmiri characters and diacritics, ensuring explicit modeling of fine-grained vowel distinctions during alignment and synthesis. To this end, our contributions are threefold:

\begin{itemize}
    \item \textbf{Bolbosh Framework:} To the best of our knowledge, we introduce the first script-aware, flow-matching-based Kashmiri TTS system.
    
    \item \textbf{Acoustic Domain Integration and Cross-Lingual Adaptation:} We develop a principled enhancement pipeline and supervised fine-tuning strategy that unify heterogeneous speech corpora and leverage a pretrained English multi-speaker checkpoint for stable low-resource adaptation.
    
    \item \textbf{State-of-the-Art Performance:} Bolbosh achieves a MOS of 3.63 and an MCD of 3.73, substantially outperforming multilingual baselines (MOS 1.86) and establishing a new benchmark for Kashmiri speech synthesis.
\end{itemize}

Beyond Kashmiri, our findings highlight a broader limitation of multilingual TTS systems in modeling diacritic-sensitive scripts and demonstrate the importance of script-aware encoding for scalable low-resource speech synthesis.

\section{Related Work}

\subsection{Speech Technology for Indic Languages}

Recent progress in Indic speech technology, driven by initiatives such as AI4Bharat and IndicParler~\cite{sankar25_interspeech}, has established multilingual TTS and ASR baselines for high-resource languages including Hindi, Tamil, Telugu, and Marathi. However, Kashmiri remains largely unsupported. While limited work has explored Kashmiri ASR and translation, no publicly available neural TTS system exists for the language. Multilingual systems that unofficially support Kashmiri perform poorly, as they are not explicitly adapted to its phonotactics or diacritic-rich Perso-Arabic script, revealing a clear need for a dedicated, script-aware TTS framework~\cite{lone2022natural, kachru2016kashmiri, koul2004kashmiri, koul1987spoken}.
Neural Text-to-Speech has evolved across several architectural paradigms, each with trade-offs that become pronounced in low-resource settings. \textbf{Autoregressive (AR)} models such as Tacotron 2~\cite{shen2018naturalttssynthesisconditioning} generate natural speech but suffer from unstable attention and alignment failures when trained on limited data. \textbf{Non-Autoregressive (NAR)} models like FastSpeech 2~\cite{ren2022fastspeech2fasthighquality} improve inference speed but require accurate duration supervision via external aligners such as the Montreal Forced Aligner~\cite{mcauliffe2017montreal}, which depends on reliable G2P resources unavailable for Kashmiri. \textbf{GAN-based} models, notably VITS~\cite{kim2021conditionalvariationalautoencoderadversarial}, remove external alignment dependencies through Monotonic Alignment Search~\cite{kim2020glow}, yet remain sensitive to hyperparameters and prone to instability in low-resource conditions. \textbf{Diffusion-based} models such as Grad-TTS~\cite{popov2021gradttsdiffusionprobabilisticmodel} provide stable training and high fidelity but require computationally expensive iterative denoising at inference.

\subsection{Flow-Matching Based TTS}

Recent advances in continuous normalizing flows \cite{lipman2022flow} and Optimal Transport Conditional Flow Matching (OT-CFM) \cite{tong2023conditional} provide a compelling alternative. Unlike diffusion models, OT-CFM learns a direct continuous-time vector field that transports a simple prior distribution to the target acoustic distribution. This formulation avoids iterative Markov chains while preserving stable training dynamics.
Flow-matching models combine several desirable properties for low-resource TTS: like stable and sample-efficient training, faster inference compared to diffusion, reduced risk of mode collapse compared to GANs, end-to-end alignment without external phonetic supervision.
We adopt \textbf{Matcha-TTS} \cite{mehta2024matcha}, a state-of-the-art OT-CFM architecture, as the backbone of our Kashmiri system. Matcha-TTS integrates Monotonic Alignment Search internally, removing the dependency on external aligners while maintaining strong alignment stability. Importantly, it operates directly on grapheme-level inputs, making it well-suited for Kashmiri, where phonetic lexicons and standardized G2P resources are scarce.
By leveraging OT-CFM within a supervised cross-lingual adaptation framework, our work demonstrates that flow-matching architectures provide a practical and scalable pathway for high-quality speech synthesis in severely under-resourced languages.

%% file: sections/method.tex
\section{Kashmiri TTS Dataset}

 We curate a 79.9-hour Kashmiri corpus, split into training, validation, and test sets (Table~\ref{tab:dataset_stats}). The training set combines studio-quality RASA recordings with the multi-speaker IndicVoices-R corpus, while validation and test splits are drawn exclusively from RASA to ensure controlled evaluation.

\subsection{Dataset}
\label{dataset}

\begin{table}[t]
\centering
\caption{Duration and utterance statistics for the Kashmiri TTS dataset. 
IndicVoices-R was used exclusively during training for acoustic prior modeling.}
\label{tab:dataset_stats}

\resizebox{\columnwidth}{!}{
\begin{tabular}{l c c c}
\toprule
\rowcolor{blue!20}
\textbf{Split} & \textbf{RASA} & \textbf{IndicVoices-R} & \textbf{Total Duration} \\
\midrule
\rowcolor{gray!8}
\textbf{Train} & 25.49 h (15,898) & 43.61 h (17,284) & 69.10 h \\
\textbf{Validation} & 7.23 h (4,542) & 0.00 h (0) & 7.23 h \\
\rowcolor{green!8}
\textbf{Test} & 3.56 h (2,272) & 0.00 h (0) & 3.56 h \\
\midrule
\rowcolor{red!15}
\textbf{Total} & \textbf{36.28 h} & \textbf{43.61 h} & \textbf{79.89 h} \\
\bottomrule
\end{tabular}
}
\end{table}

\textbf{RASA Kashmiri Dataset}~\cite{ai4bharat2024rasa} is a studio-recorded corpus containing clean, high-quality speech captured under controlled conditions. It serves as the primary acoustic backbone of our system, ensuring stable phonetic alignments and consistent pronunciation in the Perso-Arabic script.

\noindent\textbf{IndicVoices-R (Kashmiri split)}~\cite{sankar2024indicvoicesr} is a large multilingual corpus with predominantly spontaneous recordings (93.25\%) collected across diverse environments. While it provides valuable speaker and prosodic diversity, the data exhibits noise, reverberation, and amplitude variation, which can destabilize alignment and degrade synthesis quality. To address this, we apply a dedicated acoustic enhancement pipeline prior to training.

\subsubsection{IndicVoices-R Enhancement Pipeline}
\label{audioenhance}

Flow-based architectures using Monotonic Alignment Search (MAS) require clean and temporally consistent audio for stable alignment. Direct training on spontaneous multi-environment recordings can introduce alignment errors and acoustic artifacts. To bridge the gap between IndicVoices-R and the studio-quality RASA corpus, we apply a three-stage enhancement pipeline: (i) dereverberation and denoising using the Resemble-Enhance framework~\cite{resemble_enhance} with a UNet-based denoiser~\cite{ronneberger2015u} and latent CFM refinement for spectral clarity; (ii) dynamic silence trimming by removing segments below 40\,dB of peak amplitude to prevent erroneous duration assignments; and (iii) loudness normalization to $-23.0$ LUFS (ITU-R BS.1770-4~\cite{itu2015bs1770}) followed by resampling to 22.05\,kHz. This pipeline enables stable integration of spontaneous and studio recordings while preserving alignment reliability and acoustic fidelity.

\subsection{Text Processing and Normalization}

We adopt the Perso-Arabic script and design a custom normalization pipeline to address orthographic variability and diacritic sensitivity. The pipeline performs canonicalization of Unicode variants, number expansion into Kashmiri text, and character filtering while strictly preserving pronunciation-critical diacritics. Unlike conventional TTS systems, we omit explicit Grapheme-to-Phoneme (G2P) conversion, as Kashmiri exhibits moderate grapheme–phoneme correspondence when diacritics are retained. Instead, we extended model's vocabulary to 272 letters and disable language-specific text cleaners, allowing the Matcha-TTS encoder to learn end-to-end grapheme-to-acoustic mappings while preserving fine-grained vowel distinctions.

\section{Proposed Framework}

\subsection{Model Architecture}

Our Kashmiri Text-to-Speech system is built upon Matcha-TTS~\cite{mehta2024matcha}, 
extending its Optimal Transport Conditional Flow Matching (OT-CFM) formulation for script-aware low-resource cross-lingual adaptation. We fine-tune a pretrained multi-speaker English Matcha-TTS checkpoint to leverage cross-lingual acoustic priors in a low-resource adaptation setting.
The model consists of a text encoder, duration predictor, pitch and energy predictors, an OT-CFM-based decoder, and a neural vocoder. The text encoder maps normalized grapheme sequences into contextualized representations using stacked Transformer layers~\cite{vaswani2017attention}. Unlike autoregressive attention-based models, alignment is handled explicitly through a duration predictor, which estimates grapheme-level durations and enables deterministic length regulation.
Prosodic features are modeled using pitch and energy predictors operating at the grapheme level. Predicted durations expand these features to frame-level representations, which condition the acoustic decoder. The OT-CFM decoder learns a continuous velocity field that transports a simple Gaussian prior to the target mel-spectrogram distribution conditioned on linguistic and prosodic embeddings. This formulation preserves the stability advantages of diffusion-based approaches while requiring significantly fewer inference steps.
Waveform reconstruction is performed using a pretrained HiFi-GAN vocoder~\cite{kong2020hifigangenerativeadversarialnetworks}, which remains frozen during fine-tuning to ensure stable and high-fidelity synthesis.

\subsection{Training Objective}

The overall training objective combines acoustic reconstruction with auxiliary supervision:

\begin{equation}
\mathcal{L}_{total} = 
\mathcal{L}_{mel} 
+ \lambda_{dur}\mathcal{L}_{dur} 
+ \lambda_{pitch}\mathcal{L}_{pitch} 
+ \lambda_{energy}\mathcal{L}_{energy}.
\end{equation}

\noindent\textbf{Mel-spectrogram loss} $\mathcal{L}_{mel}$ supervises the OT-CFM decoder using $L_1$ reconstruction between predicted and ground-truth mel features.
\noindent\textbf{Duration loss} $\mathcal{L}_{dur}$ is computed as mean squared error between predicted and MAS-derived grapheme durations.
\noindent\textbf{Pitch} and \textbf{energy losses} are similarly defined as regression objectives at the grapheme level.
The flow-matching objective implicitly regularizes the learned velocity field to match the optimal transport trajectory between the Gaussian prior and target mel distribution~\cite{tong2023conditional}. This stabilizes training without adversarial optimization or iterative denoising.

\subsection{Low-Resource Adaptation Strategy}
\label{ssec:low_resource}

To address limited supervised data, we adopt a two-stage adaptation strategy.

\noindent\textbf{Cross-Lingual Initialization.} We initialize from a pretrained English multi-speaker Matcha-TTS model rather than training from scratch. Since flow-based decoders learn transferable acoustic representations~\cite{mehta2024matcha}, this cross-lingual initialization provides a strong prior and accelerates Monotonic Alignment Search (MAS) convergence on Kashmiri data.  Although Matcha-TTS operates at the grapheme level, the pretrained model does not include Kashmiri-specific characters and diacritics.
We therefore augment the embedding vocabulary with additional Kashmiri grapheme symbols, expanding it to 272 graphemes.

\noindent\textbf{Multi-Speaker Regularization.} To prevent overfitting to the studio-quality RASA corpus, we incorporate enhanced IndicVoices-R data during training. Each utterance is assigned a learned speaker embedding, promoting generalization across acoustic conditions. IndicVoices-R provides phonetic and prosodic diversity, while RASA anchors high-fidelity synthesis. During inference, we condition exclusively on RASA speaker embeddings to maintain studio-level clarity.

%% file: sections/results.tex
\section{Results and Discussion}

\subsection{Implementation Details}

We fine-tuned the pretrained multi-speaker Matcha-TTS model on the curated Kashmiri corpus using alignment-aware configurations tailored for low-resource adaptation. Training was conducted on a single NVIDIA H100 NVL GPU with mixed-precision (\texttt{fp16}) and eight data loader workers to ensure efficient throughput.
Optimization used Adam~\cite{kingma2014adam} with an initial learning rate of $1 \times 10^{-4}$ and no weight decay. Gradient clipping (max norm = 5.0) was applied to stabilize early MAS convergence. An effective batch size of 128 was achieved via a per-device batch size of 64 with gradient accumulation over two steps. The best checkpoint was selected based on validation loss.
Audio was standardized to 22.05 kHz and converted into 80-dimensional Mel-spectrograms using STFT parameters: FFT size 1024, window length 1024, hop length 256, and frequency range 0–8000 Hz. Mel features were normalized using dataset-specific statistics (mean $-5.603$, std $2.571$). Text input employed a grapheme-level vocabulary, explicitly preserving Kashmiri diacritics for accurate encoder conditioning.

\begin{table}[t]
\centering
\caption{\textbf{Benchmarking ASR Models for Kashmiri Proxy Evaluation.} 
We compare WER across architectures and normalization strategies.}
\label{tab:asr_benchmark}
\setlength{\tabcolsep}{3pt}

\begin{tabular}{p{2.8cm} p{1.9cm} p{2.2cm} c}
\toprule
\rowcolor{gray!20}
\textbf{Model Family} & \textbf{Model} & \textbf{Condition} & \textbf{WER} \\
\midrule

\textbf{IndicConformer \cite{indicconformer_ks_2024}} & RNN-T & With Diacritics & \cellcolor{red!15}66.59 \\
(AI4Bharat) & \textbf{RNN-T} & \textbf{No Diacritics} & \cellcolor{green!20}\textbf{41.20} \\
 & CTC & No Diacritics & \cellcolor{blue!15}44.00 \\
\midrule

\textbf{OmniASR \cite{omnilingualasr2025}} & CTC (300M) & With Diacritics & \cellcolor{red!25}94.34 \\
(Meta) & CTC (300M) & No Diacritics & \cellcolor{red!20}90.06 \\
 & LLM (7B) & No Diacritics & \cellcolor{red!18}87.67 \\
\bottomrule
\end{tabular}
\vspace{-0.5 cm}
\end{table}
\subsection{Evaluation Metrics}

We assess synthesis quality using both objective and subjective measures. 
Objective fidelity is evaluated with Mel-Cepstral Distortion (MCD)~\cite{407206}. 
To account for speaking-rate differences, synthesized and reference utterances are aligned using Dynamic Time Warping (DTW). Mel-Generalized Cepstral Coefficients (MCEPs) are extracted via the WORLD vocoder, excluding the $0^{\text{th}}$ coefficient, and MCD is computed as the scaled Euclidean distance along the optimal alignment path, where lower values indicate greater spectral similarity.
Subjective quality is measured through Mean Opinion Score (MOS)~\cite{itut1996p800}. 
We conducted a listening study with 32 native Kashmiri speakers who rated intelligibility and prosodic naturalness on a 5-point scale (1: unintelligible, 5: perfectly natural and intelligible).
We additionally report Word Error Rate (WER) using a proxy ASR system 
(\textit{indicconformer\_stt\_ks\_hybrid\_ctc\_rnnt\_large}~\cite{indicconformer_ks_2024}). To separate synthesis errors from inherent ASR limitations, we compute Relative WER (rWER) normalized against ASR performance on ground-truth recordings. Given the high baseline ASR error rate, WER is treated as a supplementary metric.

\begin{table}[t]
\centering
\caption{TTS Benchmark Results. MCD, rWER, and WER ($\downarrow$) lower is better.}
\label{tab:tts_benchmark}
\resizebox{\columnwidth}{!}{%
\begin{tabular}{l l c c c}
\toprule
\rowcolor{gray!20}
\textbf{Model} & \textbf{Condition} & \textbf{MCD} & \textbf{rWER (\%)} & \textbf{WER} \\
\midrule

\textbf{Bolbosh} 
& With Diacritics 
& \cellcolor{green!20}\textbf{3.73} 
& \cellcolor{green!20}\textbf{4.14} 
& \cellcolor{green!20}0.6935 \\

& No Diacritics 
& -- 
& \cellcolor{blue!15}13.23 
& \cellcolor{blue!15}0.4665 \\

\midrule

\textbf{IndicParler} 
& With Diacritics 
& \cellcolor{red!20}4.73 
& \cellcolor{red!20}46.75 
& \cellcolor{red!20}0.9772 \\

& No Diacritics 
& -- 
& \cellcolor{red!30}100.32 
& \cellcolor{red!30}0.8253 \\

\bottomrule
\end{tabular}%
}
\end{table}

\begin{table}[t]
\centering
\caption{Mean Opinion Score (MOS) Results for Kashmiri TTS Systems with 95\% Confidence Intervals.}
\label{tab:mos_results}
\begin{tabular}{l c c}
\toprule
\rowcolor{gray!20}
\textbf{System} & \textbf{MOS ($\uparrow$)} & \textbf{95\% CI} \\
\midrule

Human (Ground Truth) 
& \cellcolor{green!25}\textbf{4.614} 
& $\pm$ 0.059 \\

\textbf{Bolbosh (Ours)} 
& \cellcolor{blue!20}\textbf{3.634} 
& $\pm$ 0.061 \\

IndicParler (Baseline) 
& \cellcolor{red!25}1.864 
& $\pm$ 0.065 \\

\bottomrule
\end{tabular}
\vspace{-0.5 cm}
\end{table}
\subsection{ASR Proxy Evaluation}

Table~\ref{tab:asr_benchmark} compares Kashmiri ASR systems used for proxy intelligibility evaluation. The IndicConformer RNN-T model without diacritics achieves the lowest WER (41.20\%), outperforming its CTC variant (44.00\%) and all OmniASR models. Retaining diacritics substantially increases error (66.59\%), reflecting current ASR limitations in modeling diacritic-rich orthography.
Based on these results, we adopt the diacritics-removed RNN-T configuration for rWER computation. The high absolute WER across systems further indicates that Kashmiri ASR remains underdeveloped, supporting our decision to treat WER as a supplementary metric.

\subsection{Objective TTS Evaluation}

Objective synthesis performance is summarized in Table~\ref{tab:tts_benchmark}. Our proposed Bolbosh framework achieves an MCD of \textbf{3.73}, outperforming the multilingual IndicParler baseline (4.73). This represents a substantial reduction in spectral distortion, indicating significantly improved phonetic fidelity and timbral consistency.
Relative WER (rWER) further corroborates these findings. Under diacritic-preserving input conditions, our model achieves an rWER of \textbf{4.14\%}, compared to 46.75\% for IndicParler. When diacritics are removed, performance degrades (13.23\%), demonstrating that explicit diacritic modeling materially improves intelligibility. In contrast, IndicParler exhibits extreme degradation without diacritics (100.32\%), suggesting unstable grapheme-to-acoustic mapping.
These results validate two key design choices: (i) explicit diacritic modeling and (ii) supervised cross-lingual fine-tuning. The large performance margin confirms that zero-shot multilingual transfer is insufficient for Kashmiri synthesis.

\subsection{Subjective Human Evaluation}

Subjective listening results are presented in Table~\ref{tab:mos_results}. Our model achieves a MOS of \textbf{3.634} ($\pm 0.061$), significantly outperforming IndicParler (1.864 $\pm 0.065$). The near two-point improvement indicates substantial gains in intelligibility and prosodic naturalness.
While a gap remains between synthesized speech and ground-truth recordings (4.614 $\pm 0.059$), the relatively narrow confidence interval demonstrates consistent listener agreement. Notably, the baseline system frequently produced unintelligible or prosodically distorted outputs, reflected in its low MOS distribution.
These findings demonstrate that flow-matching-based supervised adaptation can elevate Kashmiri TTS from largely unintelligible zero-shot synthesis to near-natural speech quality.

\subsection{Spectral Analysis}

\begin{figure}[t]
    \centering
    \includegraphics[width=\columnwidth]{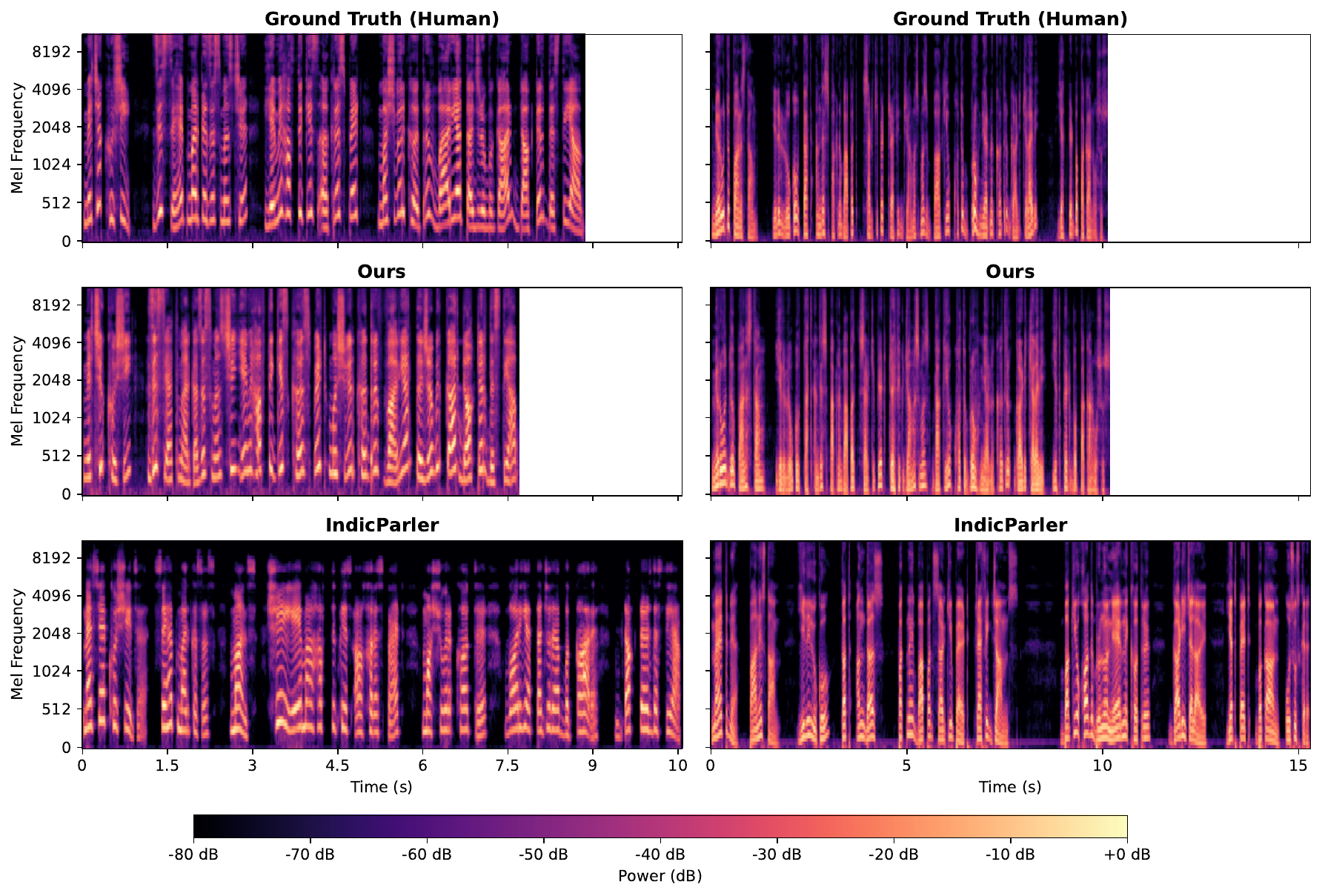}
    \caption{Mel-spectrogram comparison of 2 Kashmiri utterances. 
    The utterances were synthesized for the following text:\newline
    Right (IPA): [\textipa{me ru:\textsubbridge{d} n1 pA:n2s \textsubbridge{t}A:m jel1 lUkow \textsubbridge{t}2\textsubbridge{t}\super{h} @n\textsubbridge{d}2s p2kn1 bA:p@\textsubbridge{t}\super{h} s2\textrtaild{}kI pj2\textrtailt{}\super{h} tr\ae{}p\super{h}Ik \textdyoghlig{}A:m@s m2nz bA:kIjow k\super{h}o\textsubbridge{t}1 br\~{o}h nj@:rn1 k\super{h}@:\textsubbridge{t}r1 g@:\super{j}\textrtaild{} \textteshlig{}2l@:w j2\textsubbridge{t}\super{h} pj2\textrtailt{}\super{h} b1 p@kA:n o:sUs}]\newline
    Left (IPA): [\textipa{me \textteshlig{}\super{h}1 k\super{h}oSi: zI seh2\textsubbridge{t} m2rk@z@s m2nz j1m h2p\super{h}\textsubbridge{t}1 kIs @:k\super{h}r@s pj2\textrtailt{}\super{h} m2Sw@r1 k\super{h}@:\textsubbridge{t}r1 wA:rj@h \textsubbridge{t}2\textdyoghlig{}rUb1 kA:r \textrtaild{}A:k\super{h}\textrtailt{}@r \textsubbridge{d}2s\textsubbridge{t}IjA:b}]\newline}
    \label{fig:spectrogram_comparison}
    \vspace{-1 cm}
\end{figure}

Figure~\ref{fig:spectrogram_comparison} presents a qualitative spectrogram comparison. Bolbosh preserves clear harmonic structures and well-defined formant trajectories, with coherent high-frequency energy and stable temporal transitions. In contrast, IndicParler exhibits over-smoothing, blurred formants, and temporal instability, consistent with its higher MCD and lower MOS scores. These results further validate Bolbosh as a strong benchmark for Kashmiri TTS and demonstrate its effectiveness for low-resource speech synthesis.


%% file: sections/conclusion.tex
\section{Conclusion}

In this work, we introduced \textbf{Bolbosh}, the first dedicated, open-source Text-to-Speech system optimized for Kashmiri. We showed that zero-shot multilingual baselines fail to produce intelligible and acoustically faithful speech due to inadequate modeling of Perso-Arabic diacritics and language-specific phonotactics.
By initializing from a pretrained English multi-speaker checkpoint, 
augmenting the input embeddings with Kashmiri characters and diacritics, and employing structured multi-speaker regularization, we achieve stable alignment and high-fidelity synthesis in a low-resource setting. \textbf{Bolbosh} attains an MCD of 3.73 and a MOS of 3.63, substantially outperforming multilingual baselines and establishing a new benchmark for Kashmiri TTS.
More broadly, Bolbosh demonstrates the importance of script-aware modeling and flow-based supervised adaptation for scalable low-resource speech synthesis. Future work will explore multi-dialect modeling, enhanced prosody control, and cross-lingual transfer to other under-resourced languages.